\documentclass[10pt,conference]{IEEEtran}
\IEEEoverridecommandlockouts
\usepackage{eqparbox}
\usepackage{csquotes}
\usepackage{cite}
\usepackage{amsmath,amssymb,amsfonts}
\usepackage{multicol,multirow}
\usepackage{graphicx}
\usepackage{textcomp}
\usepackage{xcolor}
\usepackage{hyperref}
\usepackage{subfigure}
\usepackage{subcaption}
\usepackage{graphicx}
\usepackage{varwidth}
\PassOptionsToPackage{hyphens}{url}
\usepackage{tikz}
\usetikzlibrary{matrix,backgrounds}
\usepackage[printonlyused,nohyperlinks]{acronym}
\usepackage{draftwatermark}
\usepackage{array}
\usepackage{float}
\usepackage{placeins}
\usepackage{amsmath}
\usepackage{algorithm}
\usepackage{algorithmic}
\SetWatermarkText{}
\SetWatermarkScale{0.7}
\begingroup
\endgroup

\def\BibTeX{{\rm B\kern-.05em{\sc i\kern-.025em b}\kern-.08em
    T\kern-.1667em\lower.7ex\hbox{E}\kern-.125emX}}

 \usepackage[pscoord]{eso-pic}

\begin{document}

\makeatletter
    \newcommand{\linebreakand}{%
      \end{@IEEEauthorhalign}
      \hfill\mbox{}\par
      \mbox{}\hfill\begin{@IEEEauthorhalign}
    }
    
\setlength{\parindent}{5pt}
\setlength{\parskip}{0.5pt}
\setlength{\textfloatsep}{0pt}
\setlength{\dbltextfloatsep}{0pt}
\setlength{\fboxsep}{0pt}
\setlength{\tabcolsep}{5pt}

\thispagestyle{plain}
\pagestyle{plain}
\makeatother

\title{Massimo: Public Queue Monitoring and Management using Mass-Spring Model
}

\author{
\IEEEauthorblockN{Abhijeet~Kumar$^{}$\IEEEauthorrefmark{1}}
\IEEEauthorblockA{abhijeet03143@gmail.com\\KIIT University\\ Amygdala AI\IEEEauthorrefmark{2}}
\and
\IEEEauthorblockN{Unnati~Singh$^{}$\IEEEauthorrefmark{1}}
\IEEEauthorblockA{singhunnati717@gmail.com\\KIIT University\\ Amygdala AI\IEEEauthorrefmark{2}}
\and
\IEEEauthorblockN{Rajdeep~Chatterjee$^{}$}
\IEEEauthorblockA{cse.rajdeep@gmail.com\\KIIT University\\ Amygdala AI\IEEEauthorrefmark{2}}
\and
\IEEEauthorblockN{Tathagata~Bandyopadhyay$^{}$}
\IEEEauthorblockA{gata.tatha14@gmail.com\\Technical University of Munich\\ Amygdala AI\IEEEauthorrefmark{2}}
%\\
\thanks{\IEEEauthorrefmark{1} Equal contribution}
\thanks{\IEEEauthorrefmark{2}Amygdala AI, is an international volunteer-run research group that advocates for \textit{AI for a better tomorrow} \url{http://amygdalaai.org/}.}
}
\maketitle

\begin{abstract}
An efficient system of a queue control and regulation in public spaces is very important in order to avoid the traffic jams and to improve the customer satisfaction. This article offers a detailed road map based on a merger of intelligent systems and creating an efficient systems of queues in public places. Through the utilization of different technologies i.e. computer vision, machine learning algorithms, deep learning our system provide accurate information about the place is crowded or not and the necessary efforts to be taken.
\end{abstract}


\section{Introduction}
Queue management : a perennial problem \\
A challenge faced across various public settings—such as retail stores, healthcare areas, parks, and universities—is the problem of extended waiting periods and disorganized queue management. These issues negatively impact both customers and businesses by reducing employee efficiency, service quality, and income generation. Since COVID-19, the importance of effective queue management has increased significantly to ensure both customer safety and satisfaction. Traditional queue management methods rely heavily on manual observation and outdated systems, which often prove to be inaccurate and inefficient. With the rapid advancement of technology, there is a rising demand for smart systems that can efficiently manage and monitor queues.

\par
This paper introduces a new technology that utilizes advanced computer vision techniques with YOLO models (YOLOv7, YOLOv8) and machine learning algorithms, such as Linear Regression and Polynomial Regression, to develop intelligent queue management systems. By implementing a monitoring system that detects body coordinates, calculates optimal lines, and identifies outliers, the new technology aims to improve resource allocation, reduce waiting times, and enhance the overall experience for the public. The proposed system incorporates a novel approach by analyzing the tension in virtual springs connecting detected body points, and visualizes the magnitude of these forces using color-coding.

\par
The rest of the paper is organized into five sections. Related works have been discussed in Section \ref{sec2}. Section \ref{sec3} covers the necessary background concepts. The proposed method has been discussed and visualized in Section \ref{sec4}. The results and analysis are explained in Section \ref{sec5}. The algorithms are discussed in Section \ref{sec6}. The Workflow is detailed in Section \ref{sec7}. Finally, the paper has been concluded in Section \ref{sec8}.

\section{Related Work} \label{sec2}

The problem of efficient queue management, especially in public spaces, has gained significant attention in recent years, particularly due to the challenges presented by the COVID-19 pandemic. Various studies have explored the use of advanced computer vision techniques and deep learning models to address these challenges.

\subsection{Social Distancing and Crowd Management}
During the COVID-19 pandemic, the need for efficient queue management systems became critical, especially in public spaces where maintaining social distancing was paramount. Research has been conducted using YOLOv3 and Faster R-CNN models to monitor crowd density and ensure social distancing. These studies highlight the effectiveness of YOLO-based models in real-time monitoring, offering a valuable approach to managing queues in environments where physical distancing is crucial. The use of these models in detecting and monitoring crowds aligns directly with the objectives of this paper, which aims to utilize advanced versions of YOLO (YOLOv7 and YOLOv8) to develop a more sophisticated queue management system~\href{https://link.springer.com/article/10.1007/s11042-022-13718-x}{[1]}.

\subsection{Real-Time Human Detection and Counting System Using Deep Learning}
Another relevant study focuses on real-time human detection and counting systems, employing YOLOv3 and DeepSORT tracking algorithms. This research demonstrates the effectiveness of deep learning techniques in managing crowd density in public places, such as shopping malls. The implementation of such systems ensures that crowd management and social distancing measures are upheld. The insights gained from this study are particularly relevant for the development of intelligent queue management systems, as they provide a framework for using deep learning models like YOLO for real-time monitoring and resource allocation in queues~\href{https://ojs.bonviewpress.com/index.php/AIA/article/download/391/227}{[2]}.

The existing research clearly indicates the potential of YOLO models in improving queue management systems. By building on these foundational studies, this paper proposes a novel approach that leverages advanced YOLO models and machine learning algorithms to optimize queue management in various public settings.

\section{Background Concepts}
\label{sec3}

\subsection{Object Detection Models}
Object detection is a critical component in many computer vision applications, including queue management systems. YOLO (You Only Look Once) models are widely recognized for their efficiency and accuracy in real-time object detection. YOLOv7, and YOLOv8 represent successive improvements in this model family, each iteration offering enhanced performance in terms of accuracy and inference speed. YOLOv7 introduced improvements in model architecture and training techniques, while YOLOv8 further advanced these innovations with more sophisticated network designs and optimization strategies. For the purpose of this study, YOLO models are employed to detect body points in images of individuals in queues. The choice of model impacts both the accuracy of body point detection and the computational efficiency, which are crucial for real-time applications.

The following equations are used to compute precision, recall, and the F1 score:

\textbf{Precision (P)}:
\begin{equation}
P = \frac{T_P}{T_P + F_P}
\end{equation}
where \(T_P\) is the number of true positives, and \(F_P\) is the number of false positives.

\textbf{Recall (R)}:
\begin{equation}
R = \frac{T_P}{T_P + F_N}
\end{equation}
where \(T_P\) is the number of true positives, and \(F_N\) is the number of false negatives.

\textbf{F1 Score (F1)}:
\begin{equation}
F_1 = 2 \cdot \frac{P \cdot R}{P + R}
\end{equation}
where \(P\) is the precision and \(R\) is the recall.

\subsection{Regression Techniques}
Regression analysis plays a pivotal role in modeling and predicting relationships between variables. Linear regression is a fundamental statistical method used to model the linear relationship between a dependent variable and one or more independent variables. Polynomial regression extends this concept by fitting a polynomial equation to the data, which can capture more complex relationships. In the context of queue management, these regression methods are utilized to determine the optimal path that individuals should follow based on detected body points. The choice between linear and polynomial regression methods is evaluated based on their ability to accurately model the expected path and their computational efficiency.

\textbf{Linear Regression}:
\begin{equation}
y = \beta_0 + \beta_1 x + \epsilon
\end{equation}
where \(y\) is the dependent variable, \(x\) is the independent variable, \(\beta_0\) is the intercept, \(\beta_1\) is the slope, and \(\epsilon\) represents the error term.

\textbf{Polynomial Regression}:
\begin{equation}
y = \beta_0 + \beta_1 x + \beta_2 x^2 + \cdots + \beta_n x^n + \epsilon
\end{equation}
where \(y\) is the dependent variable, \(x\) is the independent variable, \(\beta_0, \beta_1, \ldots, \beta_n\) are polynomial coefficients, and \(\epsilon\) is the error term.

\subsection{Confidence Intervals}
A confidence interval provides a range of values within which the true value of a parameter is expected to lie, with a specified level of confidence. In this study, a 95\% confidence interval is employed to assess the deviation of individual body points from the predicted optimal path. Points falling outside this interval are considered outliers, indicating deviations from the expected alignment. This statistical approach is essential for identifying non-compliant individuals in the queue and assessing the effectiveness of the optimal path model.

\textbf{95\% Confidence Interval}:
\begin{equation}
\hat{y} \pm t_{n-2} \cdot SE
\end{equation}
where \(\hat{y}\) is the predicted value, \(t_{n-2}\) is the critical value from the Student’s t-distribution with \(n - 2\) degrees of freedom, and \(SE\) is the standard error of the estimate. Points falling outside this interval are considered outliers.

\subsection{Physics-Based Outlier Detection}
The concept of spring force, derived from Hooke’s Law, is applied to model interactions between individuals in a queue. According to Hooke’s Law, the force exerted by a spring is proportional to its displacement from the equilibrium position. In this study, each individual is modeled as a mass connected by springs to their neighbors. The forces acting on each individual are calculated based on the horizontal and vertical components of these interactions. Outliers are identified by comparing the calculated forces to a threshold determined using Otsu’s method, which is a technique for automatic thresholding based on image histogram analysis.

\textbf{Hooke’s Law}:
\begin{equation}
F = K \cdot x
\end{equation}
where \(F\) is the force exerted by the spring, \(K\) is the spring constant, and \(x\) is the displacement from the equilibrium position.

The horizontal and vertical components of the force vector are:

\textbf{Horizontal Component (} \(F_x\) \textbf{)}:
\begin{equation}
F_x = F \cdot \cos(\theta)
\end{equation}

\textbf{Vertical Component (} \(F_y\) \textbf{)}:
\begin{equation}
F_y = F \cdot \sin(\theta)
\end{equation}
where \(\theta\) is the angle of the force vector with the horizontal axis.

\subsection{Derivation of Spring Force}

Consider two adjacent people located at positions \( \mathbf{p}_i = (x_i, y_i) \) and \( \mathbf{p}_{i+1} = (x_{i+1}, y_{i+1}) \), and let the optimal line direction vector be \( \mathbf{e_v} = (x_n - x_1, y_n - y_1) \), where \( n \) is the total number of people in the queue.

\subsubsection*{Distance Between People}
The distance between two people is given by:
\[
d = \sqrt{(x_{i+1} - x_i)^2 + (y_{i+1} - y_i)^2}
\]

\subsubsection*{Angle Between Two Points and the Optimal Line}
The angle \( \theta \) between the line joining two people and the optimal line is:
\[
\theta = \cos^{-1} \left( \frac{(x_{i+1} - x_i)(x_n - x_1) + (y_{i+1} - y_i)(y_n - y_1)}{d \cdot \|\mathbf{e_v}\|} \right)
\]

\subsubsection*{Deformation Along the Perpendicular Direction}
The displacement due to deviation from the optimal line is:
\[
\Delta d = d \cdot (1 - \cos(\theta))
\]

\subsubsection*{Spring Force Along the Perpendicular Direction}
Using Hooke's law, the spring force along the perpendicular direction is:
\[
F_{\perp} = k \cdot \Delta d
\]
where \( k \) is the spring constant.

\subsubsection*{Components of the Force}
The force acting between two adjacent people can be decomposed into two components:
\begin{itemize}
    \item Force along the optimal line:
    \[
    F_{\parallel} = k \cdot \Delta d \cdot \cos(\theta)
    \]
    \item Force perpendicular to the optimal line:
    \[
    F_{\perp} = k \cdot \Delta d \cdot \sin(\theta)
    \]
\end{itemize}

The net force on each individual is the sum of forces from neighboring individuals. This net force helps identify outliers by detecting deviations from the optimal line in terms of force magnitude.

\subsection{Color Mapping and Visualization}
Color mapping is a powerful tool for visualizing data and enhancing interpretability. Saliency maps and colormaps are used to represent the magnitude of forces acting on individuals in the queue. The jet colormap, which spans from blue to red, is employed to visually differentiate between low, medium, and high force magnitudes. This color representation aids in quickly identifying individuals experiencing different levels of force, thus facilitating a more intuitive understanding of the queue dynamics.

\textbf{Color Mapping Function}:
\begin{equation}
\text{Color} = \text{colormap}(\text{norm\_force})
\end{equation}
where \(\text{norm\_force}\) represents the normalized force value, and the colormap function assigns colors to different ranges of force magnitudes: blue (low), green/yellow (medium), and red (high).

\section{Proposed Solution}
\label{sec4}
The proposed solution for optimizing queue management involves a multi-faceted approach integrating advanced object detection models, regression analysis, statistical methods, and visualization techniques. Each component of the solution addresses specific aspects of queue dynamics to enhance accuracy and real-time performance.

\subsection{Heap Area Coordinate Detection Methods}

To accurately detect and track individuals within a queue, we employ three distinct object detection models: YOLOv7, and YOLOv8. Each model represents an evolution in the YOLO (You Only Look Once) series, offering incremental improvements in detection capabilities and inference speed.

\begin{itemize}
    \item \textbf{YOLOv7}: This model features advanced architectural enhancements and optimization techniques, pushing the boundaries of object detection performance. It serves as a key reference point for evaluating improvements in detection accuracy and efficiency.
    
    \item \textbf{YOLOv8}: This iteration introduces more sophisticated network designs, which contribute to enhanced performance metrics. YOLOv8 is expected to demonstrate reduced inference times and improved accuracy compared to its predecessors.
\end{itemize}

For each model, we measure the inference time and assess detection accuracy using the same set of images. This comparative analysis allows us to identify the model that best balances accuracy and speed, thereby informing the choice of the most suitable model for real-time queue management.

\subsection{Finding the Optimal Line That Needs to Be Followed by Everyone}

Determining the optimal path within a queue involves analyzing the detected body points, with a focus on the hip area. The procedure comprises the following steps:

\begin{itemize}
    \item \textbf{Coordinate Averaging}: Two key points in the hip area are identified for each individual. The average coordinates of these points are computed to establish a reference position.
    
    \item \textbf{Optimal Line Calculation}: Using the averaged coordinates, we derive the equation of the optimal line that individuals are expected to follow. This is achieved through:
    \begin{itemize}
        \item \textbf{Linear Regression}: This method fits a straight line to the data, modeling the relationship between body point coordinates and the desired path. Linear regression provides a fundamental approach to determining the optimal path.
        
        \item \textbf{Polynomial Regression}: This technique extends linear regression by fitting a polynomial function to the data, accommodating more complex, non-linear relationships between coordinates and the path.
    \end{itemize}
\end{itemize}

A comparative analysis of these regression methods will be conducted to determine which approach offers the most accurate representation of the optimal line while minimizing computational requirements.

\subsection{Finding Outliers That Are Not Within the 95\% Confidence Interval of the Predicted Line}

To identify deviations from the optimal path, we calculate a 95\% confidence interval for the predicted line. This statistical approach involves:

\begin{itemize}
    \item \textbf{Confidence Interval Calculation}: We compute the 95\% confidence interval around the predicted line, establishing a range within which the true coordinates are expected to fall with 95\% confidence.
    
    \item \textbf{Outlier Detection}: Individuals whose coordinates fall outside this confidence interval are considered outliers. This method allows for systematic identification of deviations from the expected path, highlighting non-compliance within the queue.
\end{itemize}

This approach ensures that deviations are detected in a statistically rigorous manner, enhancing the accuracy of queue management.

\subsection{Finding Outliers Using the Tension in Spring Concept}

A physical-based model is employed to analyze interactions between individuals using the spring force concept:

\begin{itemize}
    \item \textbf{Spring Force Model}: Each individual is modeled as a mass connected by springs to adjacent persons. The force exerted by the springs is described by Hooke’s Law:
    \begin{equation}
    F = K \cdot x
    \end{equation}
    where \( F \) represents the force, \( K \) is the spring constant, and \( x \) denotes the displacement.
    
    \item \textbf{Force Calculation}: The horizontal and vertical components of the force vector are calculated based on the angle of the force vector.
    
    \item \textbf{Outlier Detection Using Otsu’s Method}: Otsu’s method is applied to determine a threshold for identifying significant deviations. Forces exceeding this threshold are classified as outliers. Otsu’s method optimizes the threshold value by analyzing the histogram of force magnitudes.
\end{itemize}

This physical modeling approach enhances outlier detection by incorporating force dynamics and statistical analysis.

\subsection{Showing Color Similar to Saliency Maps on Each Person According to the Magnitude of Force}

To visualize the force magnitudes affecting individuals in the queue, we utilize color mapping techniques:

\begin{itemize}
    \item \textbf{Color Mapping}: The jet colormap is applied to represent varying magnitudes of force. This colormap transitions from blue (indicating low force) to red (indicating high force):
    \begin{equation}
    \text{Color} = \text{colormap}(\text{norm\_force})
    \end{equation}
    where \(\text{norm\_force}\) is the normalized force value.
    
    \item \textbf{Color Representation}: By mapping different colors to force magnitudes, we provide an intuitive visual representation of the force distribution. This approach facilitates the identification of individuals experiencing varying levels of force, aiding in the assessment and management of queue dynamics.
\end{itemize}

Through these visualization techniques, we enhance the interpretability of the force magnitudes, contributing to more effective queue management and analysis.

\section{Results, Discussion, and Analysis}
\label{sec5}
This section presents the results obtained from implementing the proposed solution for queue management, discusses the performance of the different methods, and provides an in-depth analysis of the findings. The evaluation covers object detection models, optimal line determination, outlier detection, and visualization techniques.

\subsection{Object Detection Model Performance}

The performance of YOLOv7, and YOLOv8 was evaluated based on inference time and detection accuracy. YOLOv8 achieved the best results by considering both accuracy and inference time. Specifically:

\begin{table}[h!]
    \centering
    \renewcommand{\arraystretch}{1.5} % Adjust the row height
    \begin{tabular}{|>{\centering\arraybackslash}p{2cm}|>{\centering\arraybackslash}p{2cm}|>{\centering\arraybackslash}p{2cm}|}
        \hline
        \textbf{Model} & \textbf{Accuracy (\%)} & \textbf{Inference Time (ms)} \\
        \hline
        YOLOv7 & 94.4\% & 2.27 \\
        YOLOv8 & 97.2\% & 0.42\\
        \hline
    \end{tabular}
    \caption{Comparison of YOLO models.}
    \label{tab:yolo_comparison}
\end{table}

These results highlight YOLOv8 as the most efficient model for real-time applications in queue management.

\subsection{Optimal Line Determination}

The optimal line was determined using Linear Regression, Polynomial Regression, and Ridge Regression methods. The analysis revealed:

\begin{table}[h!]
    \centering
    \renewcommand{\arraystretch}{1.5} % Adjust the row height
    \begin{tabular}{|>{\centering\arraybackslash}p{2.8cm}|>{\centering\arraybackslash}p{3cm}|}
        \hline
        \textbf{Model} & \textbf{Accuracy (\%)} \\
        \hline
        Linear Regression & 97.47\% \\
        Polynomial Regression & 94.24\% \\
        Ridge Regression & 95.32\% \\
        \hline
    \end{tabular}
    \caption{Comparison of Regression models.}
    \label{tab:regression_comparison}
\end{table}

Linear Regression proved to be more effective in this scenario, offering a simpler and more accurate representation of the optimal line.

The accuracy is calculated using the following formula:
\[
\text{Accuracy} = \left(\frac{\text{Outliers detected by the model}}{\text{Ground truth outliers (visually identified)}}\right) \times 100
\]

\subsection{Outlier Detection Analysis}

\subsubsection{95\% Confidence Interval Method}

The 95\% confidence interval method effectively identified outliers. Outliers were marked in red, while inliers fell within the confidence range.

\begin{figure}[H]
    \centering
    \includegraphics[width=0.3\textwidth]{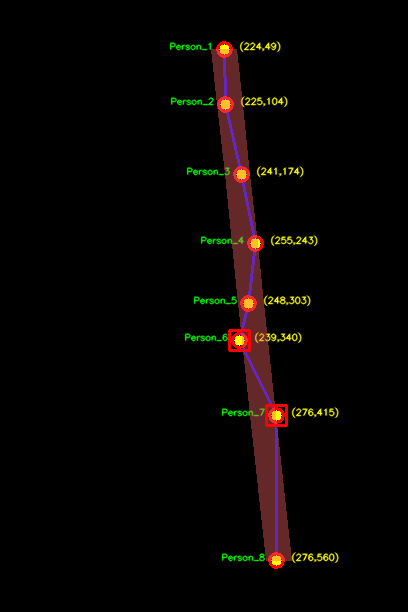} % Adjust width and path as needed
    \caption{Example of outlier detection using the 95\% confidence interval method.}
    \label{fig:confidence_interval_outlier}
\end{figure}

As illustrated in Figure 1, those who fall outside the 95\% confidence interval area are enclosed in a red rectangle, highlighting them as anomalies in the data.

\begin{figure}[H]
    \centering
    \includegraphics[width=0.3\textwidth]{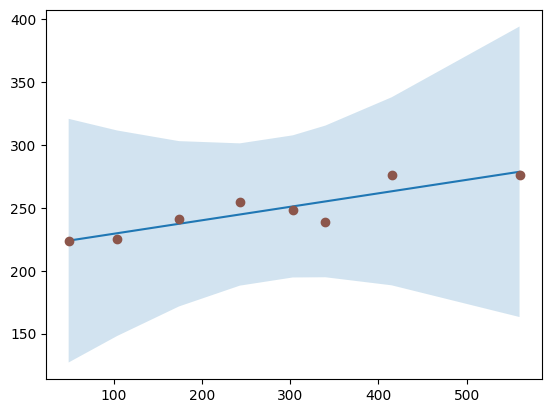}
    \caption{Graphical representation of Confidence Interval}
    \label{fig:confidence_interval_graph}
\end{figure}

\subsubsection{Spring Force Model}

The spring force model was used to detect outliers based on the tension between individuals. The method involved:

\begin{itemize}
    \item \textbf{Force Calculation}: Forces were computed using Hooke’s Law, and deviations from expected forces were assessed.
    \item \textbf{Threshold Determination}: Otsu’s method was applied to establish a threshold. Forces exceeding this threshold were considered outliers.
\end{itemize}

The spring force model demonstrated a good correlation with physical interactions and identified outliers effectively.

% \begin{figure*}[h!]
%     \centering
%     \includegraphics[width=0.9\textwidth]{tension_visualization.png} % Adjust width and path as needed
%     \caption{Visualization of spring forces between individuals.}
%     \label{fig:visualization_spring_force}
% \end{figure*}

% The first image illustrates the concept of spring forces acting between individuals along a path. Each person is connected by a spring, representing the force interaction between them. To calculate the net force acting on each person, the forces were decomposed into two components: one along the optimal line (the intended path) and the other perpendicular to it. The second image shows these components, which were then vectorially added to determine the overall force acting on each individual along both directions.

% \FloatBarrier
% \begin{figure*}[h!]
%     \centering
%     \includegraphics[width=0.5\textwidth]{tension_visualization (2).png} % Adjust width and path as needed
%     \caption{Visualization of spring forces between individuals.}
%     \label{fig:visualization_spring_force}
% \end{figure*}
% \FloatBarrier

\begin{figure}[H]
    \centering
    \includegraphics[width=0.3\textwidth]{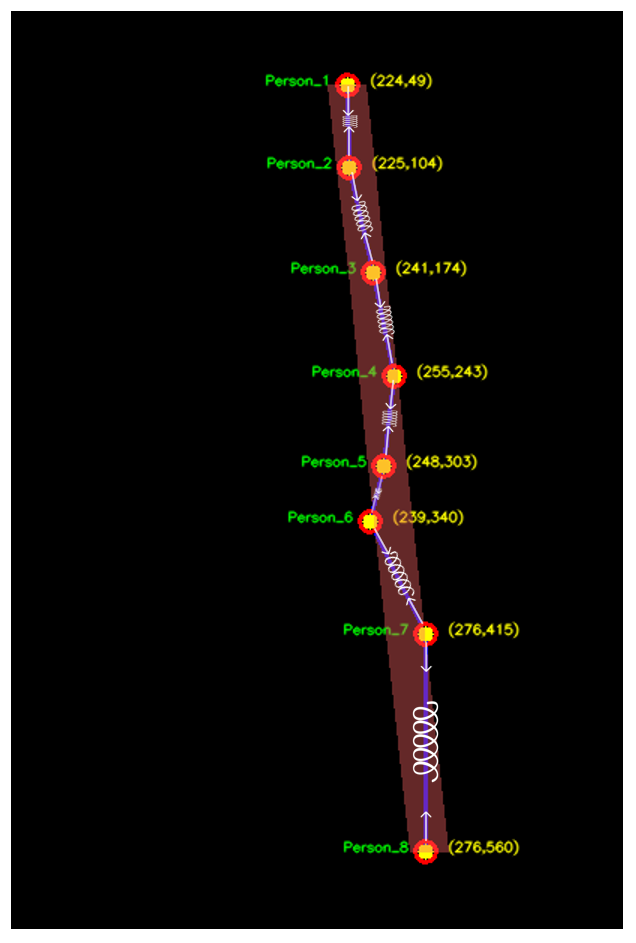} % Adjust width and path as needed 
    \caption{Visualization of spring forces acting between individuals.}
    \label{fig:visualization_spring_force1}
\end{figure}

\begin{figure}[H]
    \centering
    \includegraphics[width=0.3\textwidth]{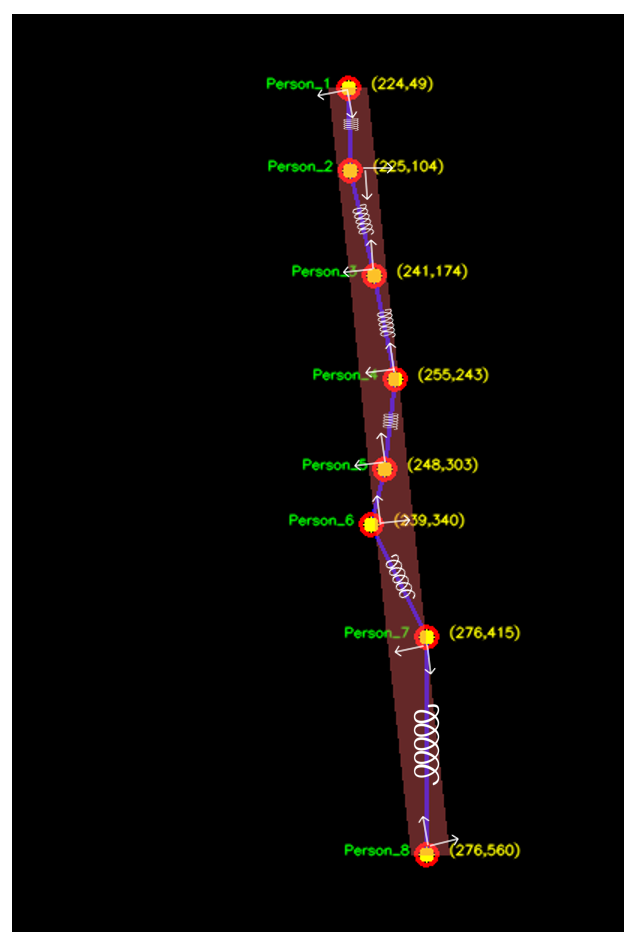} % Adjust width and path as needed
    \caption{Visualization of the components of spring forces acting between individuals.}
    \label{fig:visualization_spring_force2}
\end{figure}

Figure 3 illustrates the concept of spring forces acting between individuals along a path. Each person is connected by a spring, representing the force interaction between them. To calculate the net force acting on each person, the forces were decomposed into two components: one along the optimal line (the intended path) and the other perpendicular to it. Figure 4 shows these components, which were then vectorially added to determine the overall force acting on each individual along both directions.

\begin{table}[h!]
    \centering
    \renewcommand{\arraystretch}{1.5} % Adjust the row height
    \begin{tabular}{|>{\centering\arraybackslash}p{2.8cm}|>{\centering\arraybackslash}p{3cm}|}
        \hline
        \textbf{Model} & \textbf{No of Outliers Detected} \\
        \hline
        Regression Approach & 1 \\
        Proposed Approach & 3 \\
        \hline
    \end{tabular}
    \caption{Comparison of Regression Approach vs. Proposed Solution.}
\end{table}
\vspace{-0.2cm}
\subsection{Otsu's Method}

Otsu’s method was utilized to determine the optimal threshold for force-based outlier detection. This method involves:

\begin{itemize}
    \item \textbf{Histogram Analysis}: The method computes the histogram of the force values and divides it into two classes.
    
    \item \textbf{Variance Minimization}: Otsu's method minimizes the intra-class variance, which is the weighted sum of variances of the two classes. The intra-class variance \( \sigma^2_W(t) \) is given by:
    
    \[
    \sigma^2_W(t) = \omega_1(t) \sigma_1^2(t) + \omega_2(t) \sigma_2^2(t)
    \]
    
    where \( \omega_1(t) \) and \( \omega_2(t) \) are the probabilities of the two classes separated by the threshold \( t \), and \( \sigma_1^2(t) \) and \( \sigma_2^2(t) \) are the variances within each class.

    \item \textbf{Threshold Calculation}: The optimal threshold \( t \) is determined where the intra-class variance \( \sigma^2_W(t) \) is minimized. Forces exceeding this threshold are marked as outliers.
\end{itemize}

\textbf{Normalization and Otsu's Threshold Derivation}:

Before applying Otsu's method, we normalized the force magnitudes using min-max normalization. The equivalent force values \( \mathbf{F} \) are scaled to the range [0, 255] as follows:

\[
\mathbf{F}_{\text{scaled}} = \left( \frac{\mathbf{F} - \min(\mathbf{F})}{\max(\mathbf{F}) - \min(\mathbf{F})} \right) \times 255
\]

where \( \mathbf{F} \) is the vector of force magnitudes, \( \min(\mathbf{F}) \) and \( \max(\mathbf{F}) \) are the minimum and maximum values of \( \mathbf{F} \), respectively.

After normalization, Otsu’s threshold \( t_{\text{Otsu}} \) is calculated by maximizing the between-class variance, and forces greater than \( t_{\text{Otsu}} \) are marked as outliers.

\[
t_{\text{Otsu}} = \arg\min_t \, \sigma^2_W(t)
\]

By using Otsu's method, the threshold for force magnitudes was set in an automated and statistically sound manner, ensuring robust outlier detection.

\subsection{Visualization and Color Mapping}

\begin{figure}[h!]
    \centering
    \includegraphics[width=0.45\textwidth]{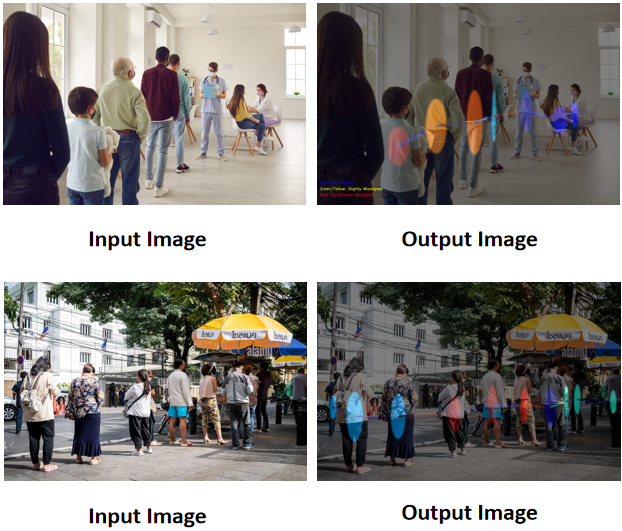} % Adjust width and path as needed
    \caption{Visualization of force magnitudes using color mapping.}
    \label{fig:visualization of colour coding}
\end{figure}

Color mapping was employed to visualize force magnitudes affecting individuals in the queue. The colormap transitioned from blue (low force) to red (high force). This visualization provides an intuitive understanding of force distribution, aiding in the assessment of queue dynamics.
\FloatBarrier
\section{Algorithms}
\label{sec6}

\begin{algorithm}
\caption{Calculating Mid Heap Area Point}
\begin{algorithmic}[1]
\STATE \textbf{Input:} Image path \( image\_path \), threshold \( th \)
\STATE \textbf{Output:} Mid heap area points of detected people
\STATE \textbf{Step 1:} Load YOLO model: 
\STATE \quad \( model \gets YOLO('yolov8n-pose.pt') \)
\STATE \textbf{Step 2:} Read the image: 
\STATE \quad \( img \gets cv2.imread(image\_path) \)
\STATE \textbf{Step 3:} Predict keypoints from the image: 
\STATE \quad \( results \gets model.predict(image\_path)[0] \)
\STATE Initialize \( heap\_midpoints \gets [] \), \( det \gets 0 \), \( concat\_df \gets \text{empty dataframe} \)

\FOR{\textbf{each} \( result \) \textbf{in} \( results \)}
    \STATE \( landmarks \gets [] \)
    \STATE Extract keypoints \( kpts \gets result.keypoints \)
    \STATE \( nk \gets \text{number of keypoints} \)

    \FOR{\( i \gets 0 \) \textbf{to} \( nk-1 \)}
        \STATE Extract \( x, y \) coordinates from \( kpts \)
        \STATE Append \( [x, y] \) to \( landmarks \)
    \ENDFOR

    \STATE \textbf{Step 4:} Create dataframe from \( landmarks \) and concatenate to \( concat\_df \)
    \STATE Calculate midpoint \( (a, b) \) between \( landmarks[11] \) and \( landmarks[12] \)
    \STATE Append midpoint \( [a, b] \) to \( heap\_midpoints \)
\ENDFOR

\STATE \textbf{Step 5:} Clean and filter non-zero points from \( heap\_midpoints \)
\STATE \( nQ \gets \text{list of valid midpoints from } heap\_midpoints \)
\end{algorithmic}
\end{algorithm}

\clearpage

\begin{algorithm}
\caption{Calculating Equivalent Force on Each Person}
\begin{algorithmic}[1]
    \STATE \textbf{Input:} Positions of individuals \( nQ \), constant \( k \)
    \STATE \textbf{Output:} Equivalent forces for each individual
    
    \STATE \textbf{Step 1:} Initialize variables
    \STATE \( numPeople \gets \text{len}(nQ) \)
    \STATE \( ev \gets \text{calculate the equivalent vector} \)
    \STATE \( forces \gets \text{initialize force array to zeros} \)

    \STATE \textbf{Define} \texttt{findMagnitude}(vector)
    \STATE \quad \text{This function calculates the magnitude of a vector.}

    \STATE \textbf{Define} \texttt{findAngle}(p1, p2, ev)
    \STATE \quad \text{This function calculates the angle between two points} \newline \text{based on the equivalent vector.}

    \STATE \textbf{Define} \texttt{findDistance}(p1, p2)
    \STATE \quad \text{This function calculates the distance between two} \newline \text{points.}

    \STATE \textbf{Step 2:} Calculate forces for each person
    \FOR{ \( i = 0 \) \textbf{to} \( numPeople - 2 \)}
        \STATE \( \theta \gets \text{findAngle}(nQ[i + 1], nQ[i], ev) \)
        \STATE \( distance \gets \text{findDistance}(nQ[i + 1], nQ[i]) \)
        \STATE \( \Delta dis \gets \text{calculate change in distance} \)

        \STATE \( forces[i][0] \gets \text{update horizontal force component} \)
        
        \IF{ \( nQ[i + 1][1] > nQ[i][1] \)}
            \STATE \( forces[i][1] \gets \text{update vertical force component} \newline \text{positively} \)
        \ELSE
            \STATE \( forces[i][1] \gets \text{update vertical force component} \newline \text{negatively} \)
        \ENDIF
        
        % \STATE \( forces[i + 1][0] \gets \text{apply equal and opposite force horizontally} \)
        % \STATE \( forces[i + 1][1] \gets \text{apply equal and opposite force vertically} \)
        \STATE \( forces[i + 1][0] \gets \text{apply equal and opposite force} \newline \text{horizontally} \)
        \STATE \( forces[i + 1][1] \gets \text{apply equal and opposite force} \newline \text{vertically} \)
    \ENDFOR

    \STATE \textbf{Step 3:} Calculate equivalent forces
    \STATE \( equivalent \gets \text{calculate equivalent forces from the force} \newline \text{array} \)
\end{algorithmic}
\end{algorithm}

\begin{algorithm}[!htb]
\caption{Coloring Individuals and Space Based on Forces}
\begin{algorithmic}[1]
    \STATE \textbf{Input:} Image path \( image\_path \), positions \( nQ \), force magnitudes \( tnet \), equivalent forces \( equivalentAlongString \)
    \STATE \textbf{Output:} Colored image
    
    \STATE \textbf{Step 1:} Load and Initialize
    \STATE Load image: \( img \gets \text{cv2.imread}(image\_path) \)
    \STATE Initialize overlay: \( overlay \gets \text{np.zeros\_like}(img) \)
    \STATE \( output \gets img.copy() \)

    \STATE \textbf{Step 2:} Normalize Forces and Magnitudes
    \STATE Normalize forces and magnitudes: 
    \STATE \( norm\_tnet \gets \text{normalize}(tnet) \)
    \STATE \( norm\_mags \gets \text{normalize}(equivalentAlongString) \)

    \STATE \textbf{Step 3:} Color Individuals and Draw Lines
    \FOR{each individual \( i \) in \( nQ \)}
        \STATE Determine color based on \( norm\_tnet[i] \) and \( norm\_mags[i] \)
        \STATE Extract coordinates: \( x, y \gets \text{getCoordinates}(i, nQ) \)
        \STATE Calculate ellipse parameters
        \STATE Draw colored ellipse on \( overlay \)

        \IF{not last individual}
            \STATE Get next coordinates
            \STATE Draw line between individuals with color based on magnitude
        \ENDIF
    \ENDFOR

    \STATE \textbf{Step 4:} Finalize Output
    \STATE Blend \( overlay \) with \( output \)
    \STATE Add labels for color ranges indicating alignment
    \STATE Save output image
\end{algorithmic}
\end{algorithm}

\vspace{-2mm}
%\clearpage
\section{Workflow for Queue Management}
\label{sec7}
This section provides a detailed overview of the queue management system’s workflow, illustrating the sequential steps from detecting body points in an input image to visualizing forces between individuals. It covers the selection of relevant points, calculation of optimal alignment, generation of a top-view for analysis, and the final computation and visualization of forces using color-coded representations based on magnitude (refer to Figure 6 for the step-by-step workflow).
% \begin{figure}[H]
%     \includegraphics[width=1\textwidth]{flowchart_first.png}
%     \captionsetup{
%         format=plain,
%         width=1\textwidth,
%         margin=2pt,
%     }
%     \label{fig:flowchart01}
% \end{figure}
% \vspace{-1.2cm}
% \begin{figure}[H]
%     \includegraphics[width=0.95\textwidth]{flowchart_colored.png}
%     \captionsetup{
%         format=plain,
%         width=1\textwidth, 
%         margin=2pt,
%     }
%     \caption{Step-by-Step Workflow of Queue Management System}
%     \label{fig:flowchart02} 
% \end{figure}
% \begin{figure}[h!]
%     \centering
%     \includegraphics[width=1\textwidth]{final_flowchart.png}
%     \caption{Step-by-Step Workflow of Queue Management.}
%     \label{fig:confidence_interval_outlier}
% \end{figure}

The process begins with an input image where all body points are detected using object detection techniques. From these detected points, heap area points are selected, or if those are not available for some individuals, the most relevant points are chosen instead. These points are used to calculate the midpoints that determine the optimal alignment line. A top-view representation is then generated to visualize this alignment. Finally, forces are calculated by treating people as masses connected by springs, and through vector addition, the resultant force on each person and the force between each pair are computed. These forces are then represented in different colors corresponding to their magnitudes.
\begin{figure*}[!htb]
    \centering
    \includegraphics[width=0.95\textwidth]{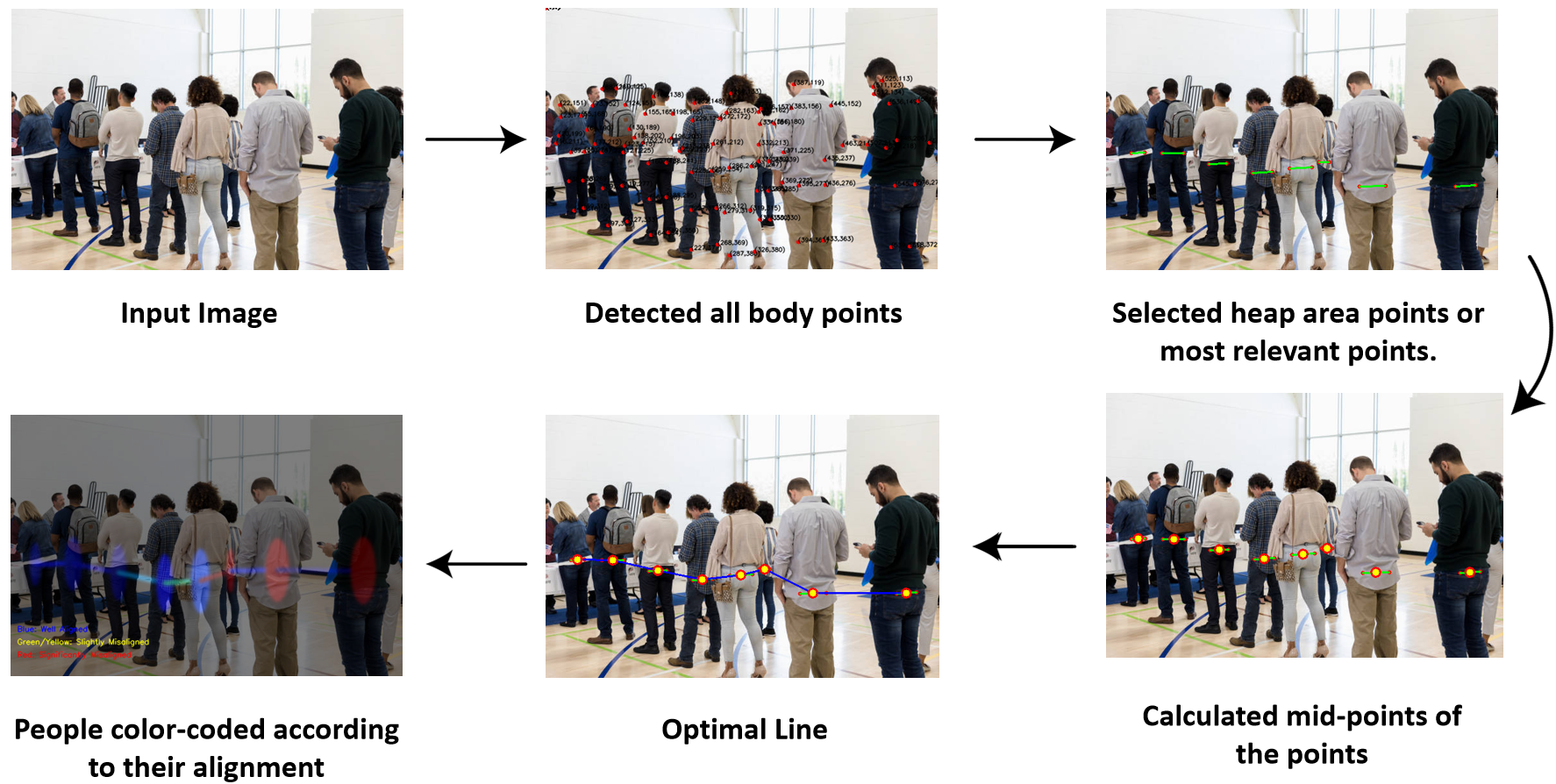}
    \centering
    \label{fig:flowchart01}
    \caption{Step-by-Step Workflow of Queue Management System}
\end{figure*}
%\clearpage
\vspace{-1mm}
\section{Conclusion}
\label{sec8}

The proposed system for public queue monitoring and management using a mass-spring model has demonstrated its effectiveness in optimizing queue management. By leveraging advanced computer vision techniques, machine learning algorithms, and statistical methods, the system provides accurate information about crowd distribution. The system successfully detects body points in frames of individuals in queues, determines the optimal line that individuals should follow, and effectively identifies outliers using two separate methods: the 95\% confidence interval method and a physics-based approach, where the spring force model is employed to analyze interactions between individuals and outliers are detected by calculating a threshold on the spring tension. Additionally, the system visualizes the magnitude of spring forces on each individual in the queue using color coding, providing an intuitive understanding of queue dynamics. With significant implications for improving resource allocation, reducing waiting times, and enhancing the overall experience for the public, the system has the potential to be integrated with other technologies and evaluated in real-world scenarios to assess its performance and identify areas for improvement.

%\vfill

\end{document}